\documentclass{bmvc2k}

\usepackage{tikz}
\usepackage{comment}
\usepackage{amsmath,amssymb} 
\usepackage{color}
\usepackage{amsmath, mathtools}
\usepackage{mathtools}
\DeclarePairedDelimiter{\abs}{\lvert}{\rvert}
\DeclarePairedDelimiter{\norm}{\lVert}{\rVert}

\usepackage{amssymb}
\usepackage{xcolor}
\usepackage{pifont}
\newcommand{\cmark}{\ding{51}}
\newcommand{\xmark}{\ding{55}}
\definecolor{forestgreen}{rgb}{0.13, 0.55, 0.13}
\definecolor{internationalorange}{rgb}{1.0, 0.31, 0.0}
\newcommand{\greencheck}{{\color{forestgreen}\cmark}}
\newcommand{\redcross}{{\color{internationalorange}\xmark}}

\usepackage[accsupp]{axessibility} 

\title{HiFECap: Monocular High-Fidelity and Expressive Capture of Human Performances}

\addauthor{Yue Jiang}{yue.jiang@aalto.fi}{1}
\addauthor{Marc Habermann \\ Vladislav Golyanik \\ Christian Theobalt}{{mhaberma,golyanik,theobalt}@mpi-inf.mpg.de}{1}
\addinstitution{
Max Planck Institute for Informatics\\
Saarland Informatics Campus
}

\runninghead{Jiang et al.}{HiFECap}

\begin{document}

\maketitle

\begin{figure}[th!]
\centering
\vspace{-24pt}
\includegraphics[width=0.8\textwidth]{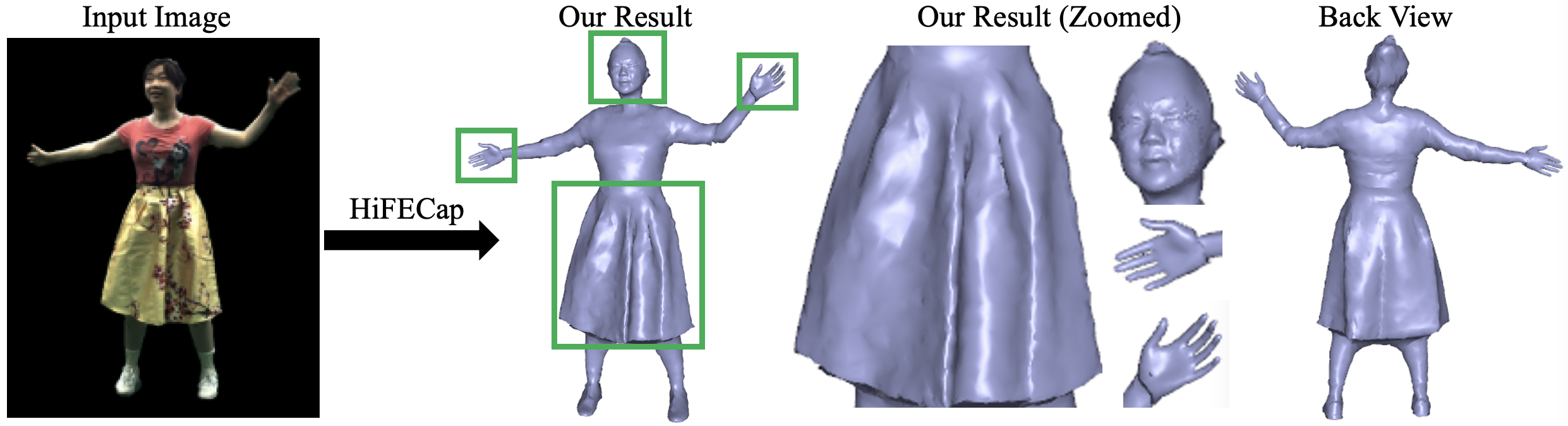}
\caption{
We present a new monocular 3D human performance capture approach, HiFECap, which for the first time jointly captures the body pose, hand gestures, facial expressions, \textit{and} high-frequency non-rigid deformations in 3D solely using RGB images as input.
The deformations recovered by our method are a clear step towards higher-fidelity cloth capture. 
\vspace{-24pt}
}
\label{fig:teaser}
\end{figure}

\begin{abstract}
Monocular 3D human performance capture is indispensable for many applications in computer graphics and vision for enabling immersive experiences. 
However, detailed capture of humans requires tracking of multiple aspects, including the skeletal pose, the dynamic surface, which includes clothing, hand gestures as well as facial expressions. 
No existing monocular method allows joint tracking of all these components. 
To this end, we propose HiFECap, a new neural human performance capture approach, which simultaneously captures human pose, clothing, facial expression, and hands just from a single RGB video. 
We demonstrate that our proposed network architecture, the carefully designed training strategy, and the tight integration of parametric face and hand models to a template mesh enable the capture of all these individual aspects.
Importantly, our method also captures high-frequency details, such as deforming wrinkles on the clothes, better than the previous works. 
Furthermore, we show that HiFECap outperforms the state-of-the-art human performance capture approaches qualitatively and quantitatively while for the first time capturing all aspects of the human.

\end{abstract}

%
%
\vspace{-12pt}
\section{Introduction} \label{sec:introduction}
%
%
The goal of 3D human performance capture is the space-time coherent tracking of the entire human surface from different sensor types; this is a long-standing and challenging computer vision problem. 
Such densely tracked characters can be used in film, game, and
mixed 
reality applications to create immersive and photo-real virtual doubles of real humans.
%
%
\par 
Previous multi-view-based approaches~\cite{bray2006posecut,brox2006high,brox2009combined,cagniart2010free,de2008performance,gall2009motion,liu2011markerless,mustafa2015general,pons2017clothcap,pons2015dyna,vlasic2008articulated,wu2013set} can capture high-quality surface details.
However, they rely on impractical and expensive multi-camera capture setups.
To commoditize performance capture, ideally, just a single RGB camera should be necessary while still allowing users to track both the body pose and non-rigid deformations of skin and clothing. 
Prior monocular approaches were able to recover the pose and shape of a naked human body model~\cite{kanazawa2018end,kanazawa2019learning,pavlakos2018learning}, hands~\cite{mueller2018ganerated,wang202rgb2hends,zhou2019monocular, chen2021camera, zhang2021hand, ge20193d, zhang2019end, boukhayma20193d}, facial expression~\cite{kim2018deep,tewari2017self,tewari17MoFA, tewari2018self, tu2019joint}, or all of them~\cite{joo2018total,pavlakos2019smplx,xiang2019monocular,zhou2021monocular}; recovering cloth deformations remains out of their reach.
Some previous work on monocular 3D human and clothes reconstruction 
uses volumetric~\cite{zheng2019DeepHuman,gabeur2019moulding} or continuous implicit representations~\cite{saito2019pifu}. 
However, these approaches do not track space-time coherent surfaces and lack surface correspondences over time. 
On the other hand, template-based monocular methods~\cite{xu18,habermann2019livecap,habermann2020deepcap,yueli21} can track low-frequency surface details coherently over time, but 
cannot capture facial expressions and hand gestures. 
Joint capture of all aspects remains poorly studied. 
%
%
\par 
To address these limitations, we present \textit{HiFECap}, a novel monocular learning-based 3D human performance capture approach that jointly captures the skeletal pose, dense surface deformations, hand gestures, and facial identity and expressions; see Fig.~\ref{fig:teaser} for an overview. 
First, convolutional neural networks predict the skeletal pose and the coarse surface deformations from the segmented monocular image of the actor. 
High-frequency surface details are recovered by a deformation network as dense vertex displacements. 
These intermediate outputs are then combined in a differentiable character representation, which can be supervised with multi-view images and 3D point clouds during training. 
We further replace the hand and face regions of the original template with parametric hand and face models using our proposed registration strategy, and drive them by predicting the parameters from images. 
%

%
%
In summary, our \textbf{technical contributions} are: 
1) HiFECap, \textit{i.e.,} the first monocular 3D human performance capture approach enabling joint tracking of body pose, the non-rigidly deforming surface, hand gestures, and facial expressions. 
2) A visibility- and rigidity-aware vertex displacement network to enable the capture of high-frequency geometric details of the dynamic human surface.
3) A multi-stage training process for surface recovery and a face and hand model integration.
Our experiments show that HiFECap applies to different clothing types and outperforms the existing state of the art in terms of recovered details.
%

%
%
\vspace{-12pt}
\section{Related Work} \label{sec:related_work}
%
%
\noindent\textbf{Multi-view Performance Capture.}  \label{sec:multiview}
Many approaches require multi-view imagery~\cite{matusik00,starck07,waschbusch05,collet15,vlasic2009dynamic}.
Prior works reconstruct surface deformations based on person-specific template  meshes~\cite{carranza03,cagniart2010free,de2008performance} or a volumetric  representation~\cite{huang16,allain15}. 
For high-quality reconstructions, some methods rely on 
segmented and high-resolution human body scans 
\cite{bray2006posecut,brox2009combined,liu2011markerless,wu12} or articulated skeletons to separate piece-wise rigid
and non-rigid deformations \cite{gall2009motion,vlasic2008articulated,liu2011markerless,wu2013set}. 
Parametric models offer another possibility to 3D human motion capture~\cite{anguelov05,hasler10,park08,pons2015dyna,loper15,kadlecek16,meekyoung17,Hesse:MICCAI:2018}.
Approaches relying on them 
often ignore clothing by treating it as noise~\cite{balan07a}, or estimate only naked body shape 
\cite{bualan08,zhang17,yang16,yinghao17}. 
Some techniques track facial expressions  \cite{joo2018total} and hands  \cite{romero2017mano,joo2018total} in addition to the proxy human body shape.  
To capture the human with clothing, some methods deform a 3D model to fit a scan~\cite{zhang17} or multi-view images~\cite{rhodin16b}, use separate  meshes for body shape and clothing~\cite{pons2017clothcap} or deploy multi-view CNNs~\cite{zeng18}. 
However, all these approaches require a  multi-view setup \textit{at inference time}, 
which makes them impractical for most users. 
In contrast, our method only leverages a multi-view setting \textit{for capturing training data}. 
Once our high-fidelity, expressive, and personalized approach is trained, it only takes a single RGB video as input at inference time. 
%
%

\noindent\textbf{Monocular 3D Pose Estimation and Performance Capture.} \label{sec:monocular}
Monocular performance capture is an ill-posed problem with lots of ambiguities (e.g., along the depth dimension and due to occlusions).
Leveraging 2D and 3D joint detections, many methods capture 3D human motion from monocular images by predicting 3D poses~\cite{popa17,zhou17,sun17,tome17,rogez17,tekin17,mehta17} or fit a parametric body model~\cite{loper15,bogo16,lassner17,kanazawa2018end,kolotouros19, varol2018bodynet, alldieck2018video}. 
Other methods directly regress the body model parameters~\cite{kanazawa2018end,pavlakos2018learning,kanazawa2019learning} or a coarse volumetric body shape~\cite{varol18}, and can also jointly capture body pose with facial expressions and hand gestures~\cite{xiang2019monocular,pavlakos2019smplx,zhou2021monocular,feng2021PIXIE}.
PIFuHD~\cite{saito2020pifuhd} and SelfRecon~\cite{jiang2022selfrecon} work for standing poses and do not generalize to arbitrary poses. 
Moreover, PIFu[HD]~
\cite{saito2019pifu, saito2020pifuhd} reconstructs per-frame geometry, while our work aims at tracking a space-time coherent geometry, which by nature is in correspondence over time. 

Capturing the non-rigid and dynamic surface of the person's clothing from monocular videos remains challenging. 
MonoClothCap~\cite{xiang20} estimates the deforming surface without the need of a person-specific template.
Instead, they deform a parametric body model during capture. 
However, they cannot track clothing types with a topology that is significantly different from the body model, e.g. skirts and dresses.
%
%
\par 
Most closely related to our work are template-based monocular 3D human performance capture methods \cite{xu18,habermann2019livecap,eventCap2020CVPR,habermann2020deepcap,yueli21}. 
MonoPerfCap~\cite{xu18} tracks an actor observed in a monocular video using a 3D actor's template. 
This method is based on global energy optimisation, and, hence, its  runtime is high and the results appear oversmoothed in many cases. 
In contrast, LiveCap~\cite{habermann2019livecap} achieves real-time performance and DeepCap~\cite{habermann2020deepcap} further improves 3D accuracy by employing a neural architecture and using multi-view supervision during training. 
Further, replacing the geometric surface regularization (e.g.,  as-rigid-as-possible regulariser) with a physics-based constraint  improves the physical plausibility of the deformations \cite{yueli21}. 
All of the above-mentioned methods 
cannot regress facial expressions, hand  gestures, and high-frequency deformations. 
In contrast, our HiFECap approach captures the state-specific appearance of the face and hands and high-frequency surface details---for the first time---in a single framework for expressive 3D human performance capture. 
%
%
%
\vspace{-12pt}
\section{Method} \label{sec:approach}
%
%
\begin{figure}[t]

\centering

\includegraphics[width=0.9\textwidth]{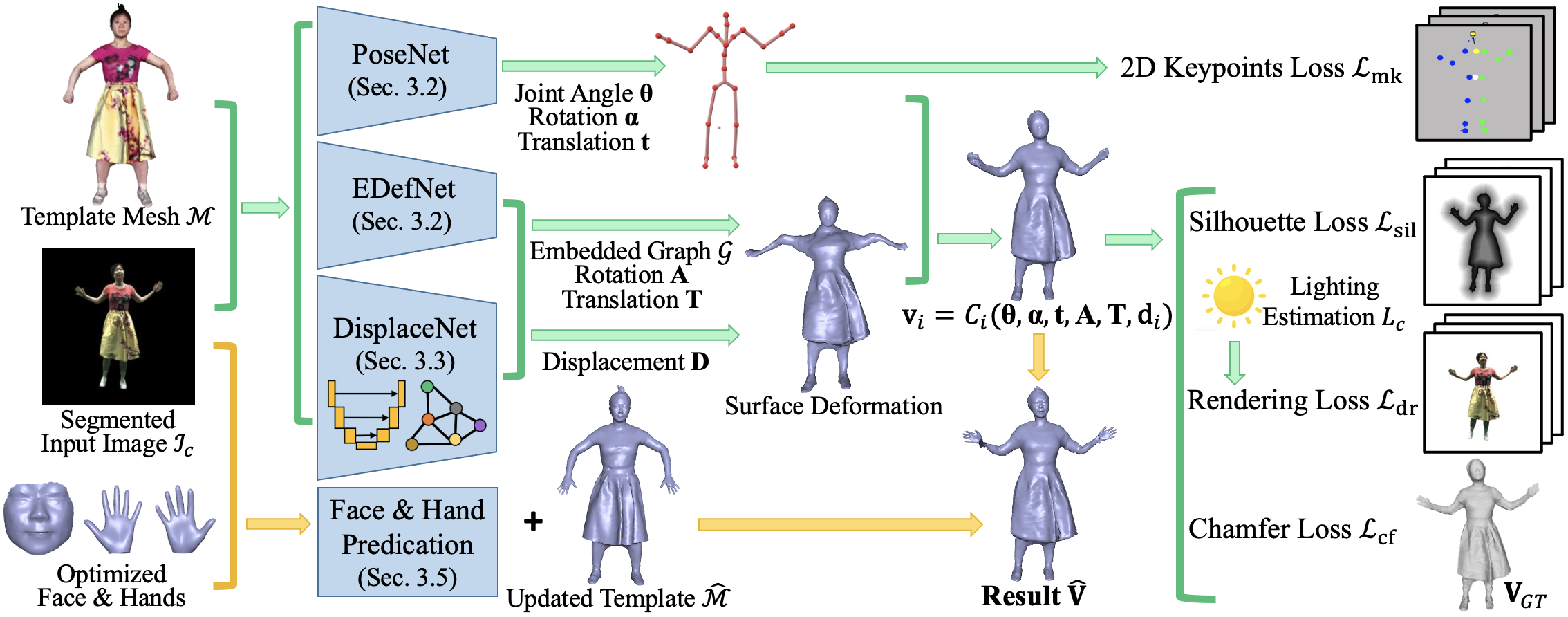}
\caption{
\textbf{Overview of our HiFECap approach} 
that takes a single segmented image as input and tracks the corresponding 3D human mesh. 
\textit{PoseNet} estimates the 3D skeletal pose as joint angles and a global rotation. 
It is followed by coarse-to-fine deformation regression based on silhouette, rendering and Chamfer losses. 
\textit{EDefNet} captures coarse 
skin and clothing details by predicting the deformation on the embedded graph.
\textit{DisplaceNet} refines the results with high-frequency details based on a vertex displacement field (green arrows).
We then replace the corresponding template parts 
with parametric hand and face models. 
Given the input image, a dedicated network then predicts those parameters (yellow arrows).
} 
\label{fig:overview}
\end{figure}

Given a monocular video of a human in motion, the goal of our method is to regress the 3D deformation of a person-specific template mesh of the human including clothing, hand gestures, and facial expressions for each of the video frames.
To this end, we first acquire multi-view training images of the human performing a diverse set of actions and define a differentiable character representation, which efficiently parameterizes the template from coarse to fine (Sec.~\ref{sec:training_data}).
Then, we propose regression networks in a coarse-to-fine manner, {\em i.e.,} we first employ a skeletal pose prediction network and a coarse embedded deformation network, which captures the piece-wise rigid skeletal deformations and the coarse surface deformations, respectively (Sec.~\ref{sec:pose_network}).
For capturing finer surface details, we propose a novel hybrid image-to-graph convolutional architecture for predicting per-vertex displacements, which greatly improves the dynamic surface details (Sec.~\ref{sec:vertex_displacement}).
Since supervised learning of the network models is not possible, we resume to a weakly-supervised setup and propose a carefully designed combination of loss functions all geared towards high fidelity surface capture (Sec.~\ref{sec:supervision}).
Last, we replace the face and hand regions of the original template with parametric models using our proposed registration procedure. 
Then, a dedicated network is predicting the facial expression as well as the hand gestures (Sec.~\ref{sec:totalcapture}). 
%
%
\subsection{Data Processing and Character Representation}\label{sec:training_data} 
For training, we record a multi-view video of the actor performing various motions in a studio with a green screen background. 
We detect 2D joint keypoints using OpenPose~\cite{cao2019openpose}, apply color keying to extract foreground masks, and generate respective distance transform images~\cite{borgefors1968distance} for each view and frame. 
We use a multi-view stereo reconstruction software Agisoft Metashape~\cite{metashape} to reconstruct the ground truth mesh $\mathbf{V}_{\mathrm{GT},f}$ for each frame. 
As input to our method, we randomly sample cropped and segmented frames $\mathcal{I}_{f, c}$ where $f$ and $c$ denote the frame and camera index. 
At the same time, other views of frame $f$ are used for supervision.
For testing, we record in-the-wild monocular videos, extract the foreground masks using Detectron2~\cite{wu2019detectron2}, and use OpenPose for retrieving 2D keypoint detections. 
For simplicity, we omit the subscript $f$ in the following. 
%
%
\par 
Our method requires a person-specific textured, rigged, and skinned 3D template of the actor. 
Therefore, we scan the person in a multi-view stereo scanner~\cite{treedys} and use Metashape~\cite{metashape} to reconstruct the 3D mesh with around $N{\approx}5000$ vertices. 
We rig the scanned mesh to a kinematic skeleton being parameterized with the root rotation $\boldsymbol{\alpha}\in\mathbb{R}^3$, the global translation $\mathbf{t}\in\mathbb{R}^3$, and the joint angles $\boldsymbol{\theta}\in\mathbb{R}^{33}$. 
We also attach 3D landmarks to the skeleton (21 body joints and 6 face landmarks). 
We automatically compute the skinning weights in Blender~\cite{blender} and leverage Dual Quaternion Skinning~\cite{kavan2007skinning} to deform the mesh based on the skeletal pose. 
Similar to Habermann et al.~\cite{habermann2019livecap}, we assign a rigidity weight $r_i$ to each vertex $\mathbf{V}_i$ to account for the different deformation properties of varying materials. 
We further define a downsampled version of the mesh as the underlying embedded graph $\mathcal{G}$ and model deformations from coarse to fine using an embedded deformation~\cite{sumner2007embedded}.
$\mathcal{G}$ is parameterized with $\mathbf{A} \in\mathbb{R}^{K\times3}$ and $\mathbf{T}\in\mathbb{R}^{K\times3}$ representing local graph rotations and translations, respectively.
Here, $K$ denotes the number of graph nodes.
To capture finer high-frequency geometric details of non-rigid deformations such as garment folds, we use a 3D vertex displacement map $\mathbf{D}$, \textit{i.e.,} we assign a displacement vector $\mathbf{d}_i\in\mathbb{R}^3$ to each mesh vertex. 
Similar to DDC~\cite{habermann21}, the final character representation $\mathbf{v}_i = C_i(\boldsymbol{\theta}, \boldsymbol{\alpha}, \mathbf{t}, \mathbf{A}, \mathbf{T}, \mathbf{d}_i)$, 
internally applies the embedded deformation and the vertex displacements in a canonical T-pose based on the parameters $\mathbf{A}, \mathbf{T}, \mathbf{d}_i$ and finally poses the deformed mesh based on the skeletal pose $\boldsymbol{\theta}, \boldsymbol{\alpha}, \mathbf{t}$.

%
%
\subsection{Pose Network and Embedded Deformation Network} \label{sec:pose_network} 
Similar to DeepCap~\cite{habermann2020deepcap}, \textit{PoseNet} and \textit{EDefNet} are both Resnet50-based networks~\cite{he2016residual} that take the image $\mathcal{I}_{c}$ of view $c$ as input.
The network architecture of \textit{EDefNet} is the same as \textit{DefNet} in DeepCap~\cite{habermann2020deepcap}, but we train it using additional loss terms in addition to the 2D supervision proposed in DeepCap (details in supplemental document).
\textit{PoseNet} regresses the skeleton joint angles $\boldsymbol{\theta}\in\mathbb{R}^{27}$ and the camera relative root rotations $\boldsymbol{\alpha}\in\mathbb{R}^3$.
The global translation of the mesh template $\mathbf{t}\in\mathbb{R}^3$ is obtained by a global alignment layer~\cite{habermann2020deepcap}.
We supervise \textit{PoseNet} with a multi-view 2D keypoint loss and a joint angle regularizer as proposed in DeepCap~\cite{habermann2020deepcap}.
\textit{EDefNet} regresses the embedded deformation parameters $\mathbf{A}, \mathbf{T}$, which capture the coarse surface deformations.
%
%
\subsection{Visibility- and Rigidity-aware Vertex Displacement Network}\label{sec:vertex_displacement}
To capture high-frequency geometric details, we add a per-vertex displacement network, \textit{DisplaceNet}, which takes the input image $\mathcal{I}_{c}$ and regresses the vertex displacement field $\mathbf{D}\in\mathbb{R}^{N\times 3}$ in the canonical pose ($\mathbf{d}_i$ denotes the $i$-th row of $\mathbf{D}$).
For this task, we found that local image patches usually contain most of the relevant information about the high-frequency deformation patterns.
Thus, we introduce a novel architecture, which maps local image features onto a graph convolutional architecture to improve accuracy and robustness.
%
%

\noindent\textbf{Image Feature Map Encoder.} 
First, we use a U-Net-based image encoder~\textit{DUNet}, which extracts relevant information about the surface deformation, {\em e.g.}, wrinkles from the input image.
More precisely, it takes the input frame $\mathcal{I}_{c}\in\mathbb{R}^{256\times256\times3}$ and computes a latent feature map $f_{\mathrm{DUNet}, c}(\mathcal{I}_{c}) = \mathcal{F}_{c}\in\mathbb{R}^{256\times256\times32}$ with the same spatial resolution as the input frame. 
%
%

\noindent\textbf{Visibility-aware Vertex Feature Map.}
Next, those features in image space are mapped onto the posed and coarsely deformed mesh: We propose a function $P_c(\mathbf{V}_i) = \mathcal{F}_{c, u, v}$ projecting the mesh vertices into image space by employing rasterization and mapping the image features at the 2D projected position $(u,v)$ to the respective graph node.
Note that the rasterization is occlusion-aware, \textit{i.e.,} only visible vertices have an attached image feature. 
However, in a monocular setting, a significant amount of vertices is usually occluded although we also want to regress their deformations.
Therefore, we argue that the whole image (of visible surface parts) can still guide the deformation state of occluded parts, e.g., the body pose can roughly give a hint of the deformations of occluded parts.
Thus, we have a visibility sensitive feature attachment function $F_c(\mathbf{V}_i) = \mathcal{F}_{c, u, v}\  \textrm{if}\ f_{\textrm{Visible}, c}(\mathbf{V}_i)$ and $F_c(\mathbf{V}_i) = a(\mathcal{F}_{c})$ otherwise,
which assigns the projected feature if a vertex is visible and for occluded vertices, it assigns the average feature, {\em i.e.,} $a(\cdot)$ averages the per-pixel features over the spatial domain.
%
%

\noindent\textbf{Visibility- and Rigidity-aware Graph Convolutional Network.}
Once each graph node has an image feature, we employ a graph CNN, 
\textit{DGCN}~\cite{habermann21} taking these per-node features and outputting the vertex displacement field 
$f_{\mathrm{DGCN}}(\textit{F}_c(\mathbf{V})) = \mathbf{D}'\in\mathbb{R}^{N\times3}$. 
Here, the graph is defined by the template mesh itself.
Nearly rigid human body parts (e.g., skin and shoes) should (if at all) only be coarsely deformed. 
Thus, we create a rigid mask $\mathbf{M} \in \mathbb{R}^{N \times 3}$ whose entries of row $i$ are set to one if $r_i \leq \epsilon_\textrm{Rigid}$ and zero otherwise.
Here, $\epsilon_\textrm{Rigid}$ is a threshold.
The displacement field is defined as $\mathbf{D} = f_{\mathrm{DGCN}}(\textit{F}_c(\mathbf{V})) \circ \mathbf{M}$, 
where $\circ$ is the Hadamard product.
%
%
\subsection{Training of EDefNet and DisplaceNet} \label{sec:supervision}
%
%
\noindent\textbf{Image-based Supervision.} 
The silhouette loss $\mathcal{L}_{\textrm{sil}}$ encourages the deformed mesh to fit the image silhouettes from all the cameras.
As such energy term can be stuck in local minima due to bad initialization, we employ a 2D multi-view landmark term $\mathcal{L}_{\textrm{mk}}$ as the difference between the projected 3D markers on the posed skeleton and the detected 2D markers on the multi-view images.
However, the aforementioned losses are not sufficient to supervise fine deformations such as surface folds.
Thus, we deploy a dense rendering loss, which takes the posed and deformed mesh and the static texture, renders it from various camera views, and compares the rendered images $R_c(\mathbf{V}, \mathbf{L}, \mathcal{T})$ under the camera's lighting condition $\mathbf{L}$ with the corresponding input frame $\mathcal{I}_c$, $\mathcal{L}_{\textrm{dr}}(\mathbf{V}, L) = \sum_c \norm{R_c(\mathbf{V}, \mathbf{L}_c, \mathcal{T}) - \mathcal{I}_c}^2$.
Assuming Lambertian surface and smooth lighting, we employ the spherical harmonics (\textit{SH}) lighting model~\cite{muller2006spherical} to represent the scene lighting $L_c(\mathbf{V}, \mathbf{l}_c)$ of each camera with $27$ coefficients $\mathbf{l}_c\in\mathbb{R}^{9\times3}$.
Then, the lighting condition for each camera can be computed as $L_c(\mathbf{V}, \mathbf{l}_c) = \sum_{j=1}^9 \mathbf{l}_{c,j}\textit{B}_j(n_c(\mathbf{V}))$ where $n_c(\mathbf{V})$ is the pixel normals of the geometry from the camera $c$.
Since scene lighting is assumed to be unknown, we also optimize it as described later.
%
%

\noindent\textbf{Chamfer Loss.}
$\mathcal{L}_{\textrm{dr}}$ helps to recover in-camera-plane deformations but 
struggles with capturing deformations along the camera viewing direction.
Therefore, we employ a Chamfer loss between the posed and deformed mesh $\mathbf{V}$ and the per-frame stereo reconstructions $\mathbf{V}_\mathrm{GT}$: $\mathcal{L}_{\textrm{cf}}(\mathbf{V}) = \sum_i \min_j\norm{\mathbf{V}_i - \mathbf{V}_\mathrm{GT,j}}^2 + \sum_j \min_i\norm{\mathbf{V}_\mathrm{GT,j} - \mathbf{V}_i}^2$.
Note that it can suffer from drifts along the surface, which  $\mathcal{L}_{\textrm{dr}}$ can prevent. 
Thus, we use the combination of two. 
%
%

\noindent\textbf{Spatial Regularization.}
To regularize the deformations, we impose an as-rigid-as-possible regularizer on the deformation graph~\cite{sorkine2007arap} and use material-aware weighting factors~\cite{habermann2020deepcap} to deal with different levels of rigidity.
We also employ a Laplacian $\mathcal{L}_{\textrm{lap}}$ and isometry $\mathcal{L}_{\textrm{iso}}$ regularization on the deformed and posed template mesh similar to Habermann et al.~\cite{habermann2019livecap}.
%
%

\noindent\textbf{Training stages.}
We train the \textit{EDefNet} in two phases while keeping the trained \textit{PoseNet} fixed. 
At this stage, the displacements are set to zero in the character representation.
We first train the embedded deformation network using the combined loss $\mathcal{L}_{\textrm{EDefNet}} = \mathcal{L}_{\textrm{sil}} + \mathcal{L}_{\textrm{mk}} + \mathcal{L}_{\textrm{arap}}$~\cite{sorkine2007arap}.
The different weights do not affect the training significantly; thus, we currently use an equal weighted sum in all experiments.
Once converged, the lighting parameters are optimized as an in-between step. 
Note that this is only possible when coarse deformations are already learned, and thus the model already roughly overlays to the ground truth images.
To optimize the lighting coefficients across all the frames in the training sequence, we minimize $\norm{R_c(\mathbf{V}, L_c(\mathbf{V}, \mathbf{l}_c), \mathcal{T}) - \mathcal{I}_c}^2$ by iteratively sampling the training frames. 
Here, $\mathcal{T}$ is the static template texture.
After convergence, we train EDefNet further and add the differentiable rendering loss~\cite{liu2019soft, jiang2020sdfdiff} $\mathcal{L}_{\textrm{dr}}$ using the optimized scene lighting and the Chamfer loss $\mathcal{L}_{\textrm{cf}}$.
%
%
\par 
While training \textit{DisplaceNet}, the weights of \textit{PoseNet} and \textit{EDefNet} are fixed and the character representation adds the displacements on top of the embedded deformation.
For supervision, we leverage the combined loss function $\mathcal{L}_{\textrm{DisplaceNet}} = \mathcal{L}_{\textrm{sil}} + \mathcal{L}_{\textrm{dr}} + \mathcal{L}_{\textrm{cf}} + \mathcal{L}_{\textrm{iso}} + \mathcal{L}_{\textrm{lap}}$.
More details regarding the losses are provided in the supplemental document.
%
%
\subsection{Tracking of Hands and Face}\label{sec:totalcapture}
So far, only the skeletal pose and the surface deformations are tracked.
To enable the joint tracking of hands and face as well, we replace the face and hand regions on the template mesh with a parametric 3D face model~\cite{blanz1999morphable} and hand model~\cite{romero2017mano} as described in the following.
%
%

\noindent\textbf{Face.}
We leverage the parametric face model of Blanz and Vetter~\cite{blanz1999morphable} with the surface geometry being defined as 
$\mathbf{V}_\mathrm{F} = 
\overline{\mathbf{V}}_{\mathrm{F}} 
+ 
\sum_{i=1}^{80}\mathbf{w}_{\textrm{S}, i} \sigma_{\textrm{S}, i} \mathbf{B}_{\textrm{S}, i} 
+ 
\sum_{j=1}^{64}\mathbf{w}_{\textrm{E}, j} \sigma_{\textrm{E}, j} \mathbf{B}_{\textrm{E}, j}$
, where $\overline{\mathbf{V}}_{\mathrm{F}}$ is the mean face.
$\mathbf{B}_{\mathrm{S}, i}\in \mathbb{R}^{80 \times 53490}$, and $\mathbf{B}_{\mathrm{E}, i}\in \mathbb{R}^{64 \times 53490}$ are the PCA bases for shape and expression variations. 
$\sigma_{\mathrm{S}, i}$ and $\sigma_{\mathrm{E},i}$ are the corresponding standard deviations. 
$\mathbf{w}_{\mathrm{S},i}\in \mathbb{R}^{80}$ and $\mathbf{w}_{\mathrm{E},i}\in \mathbb{R}^{64}$ are the face shape and expression parameters. 
We removed the neck and ear parts of the model to better fit our template in the following.
%
%

\noindent\textbf{Hands.}
We utilize the parametric MANO  model~\cite{romero2017mano} for the hands
embedded in a joint and fully articulated body and hand model called SMPL+H model defined as $\mathbf{M}_\mathrm{SH} = \overline{\mathbf{M}}_{\mathrm{SH}} + \sum_{i=1}^{16}\mathbf{w}_{\textrm{SH}, i} \mathbf{B}_{\textrm{SH}, i}$, where $\overline{\mathbf{M}}_{\mathrm{SH}}$ is the mean body shape with hands.
$\mathbf{B}_{\textrm{SH},i} \in \mathbb{R}^{16\times6890}$ are the PCA basis for body shapes with hands and $\mathbf{w}_{\textrm{SH},i}\in\mathbb{R}^{16}$ are the PCA coefficients. 
The posed and deformed mesh is defined as $\mathbf{V}_\mathrm{SH} = B(\mathbf{M}_\mathrm{SH}, \mathbf{W}, \boldsymbol{\theta}_b, \boldsymbol{\theta}_h)$ where $B(.)$ is the linear blend skinning function, $\mathbf{W}$ is the skinning weights, $\boldsymbol{\theta}_b\in\mathbb{R}^{22}$, and $\boldsymbol{\theta}_h\in\mathbb{R}^{15\times2}$ are joint angles for body and hands.
We set $\boldsymbol{\theta}_b$ and $\boldsymbol{\theta}_h$ to zero to obtain the deformed model in the canonical T-pose $\mathbf{\hat{V}}_\mathrm{SH}$ and then only consider the MANO vertices $\mathbf{\hat{V}}_\mathrm{H}$.
%
%

\noindent\textbf{Unposing to the Canonical Pose.}
The original 3D template mesh $\mathcal{M}$ in the rigging pose can differ in terms of its local rigid rotation with respect to the face and hand model. 
For a better optimization for the personalized face model and hand stitching, we unpose the original 3D template to the canonical pose $\mathcal{M}_{\mathrm{tpose}}$ using Dual Quaternion Skinning.
%
%

\noindent\textbf{Personalized Face Model.}
To retrieve the personalized face model, we fit the face model to the original template in the canonical pose $\mathcal{M}_{\mathrm{tpose}}$ as follows.
First, we optimize the affine transform between the model and the template by optimizing the affine parameters including Euler angles for rotation $\boldsymbol{\alpha}_{\mathrm{F}}\in \mathbb{R}^3$, translation vector $\mathbf{t}_{\mathrm{F}}\in \mathbb{R}^3$, and scaling $\textit{s}_{\mathrm{F}}\in \mathbb{R}$. 
We convert the Euler angles $\boldsymbol{\alpha}_{\mathrm{F}}$ to a rotation matrix $\mathbf{R}_{\mathrm{F}}\in \mathbb{R}^{3\times 3}$. 
Then, the updated face model vertices $\mathbf{V}'_{\mathrm{F}}$ can be computed as $\mathbf{V}'_{\mathrm{F}} =\textit{s}_{\mathrm{F}}\mathbf{R_{\mathrm{F}}}(\mathbf{V}_{\mathrm{F}}) +\mathbf{t}_{\mathrm{F}}$.
To optimize the affine parameters, we manually mark 8 facial landmarks on the scanned template mesh and the face model (2 on each eye, 2 on lips, 1 on nose, 1 on jaw), respectively, and minimize the difference between the two sets in the least-squares sense.
We then fix the affine transform and deform the face model to match the template geometry by optimizing the shape parameters $\mathbf{w}_{\mathrm{S}}$ and the expression parameters $\mathbf{w}_{\mathrm{E}}$. 
To this end, we minimize the Chamfer distance between the face model and template mesh as well as the distance between the two sets of markers.
Then, we once more optimize the affine parameters by minimizing the aforementioned distances.
Finally, we directly optimize the positions of the face model vertices by minimizing the Chamfer distance resulting in the updated face model position $\mathbf{V}''_{\mathrm{F}}$. 
We retrieve our final neutral face as $\hat{\mathbf{V}}_{\mathrm{F}} = \mathbf{V}''_{\mathrm{F}} - \sum_{j=1}^{64} \hat{w}_{\mathrm{E}, j} \boldsymbol{\sigma}_{\mathrm{E}, j} \mathbf{b}_{\mathrm{E}, j}$ where $\hat{w}_{\mathrm{E}, j}$ are the optimized face expression parameters.
%
%
\noindent\textbf{To connect the hand and face models with the template,} 
we use an automated gap-filling technique in Blender~\cite{blender} in the canonical space $\mathcal{\hat{M}}_{\mathrm{tpose}}$.
Finally, we repose the template to get the updated template mesh in the rigging pose $\mathcal{\hat{M}}$. 
Therefore, the skinning weights of the closest template vertex are copied to the hand and face model vertices.
%
%

\noindent\textbf{Regression of Hand and Face Parameters and Posing.}
Given an input frame $f$, we use the pre-trained model of Zhou et al.~\cite{zhou2021monocular} to regress the facial expression parameters $\mathbf{w}_{\mathrm{E, f}}$ and the hand pose parameters $\boldsymbol{\theta}_{H, f}$.
Importantly, we do not leverage the regressed face shape, but we use our optimized face model as the identity. 
We then apply the regressed facial expression and hand pose parameters to the template in the canonical pose. 
Finally, we pose and deform the template by using the regressed embedded deformation $\mathbf{A}, \mathbf{T}$ from $EDefNet$, the displacement map $\mathbf{D}$ from $DisplaceNet$, and the body pose $\boldsymbol{\theta}, \boldsymbol{\alpha}, \mathbf{t}$ from $PoseNet$.
%
%
\vspace{-12pt}
\section{Results} \label{sec:results}
%
%
\noindent\textbf{Data.}
Our method is person-specific and we aim at 
generalization to novel poses and environments.
Thus, we captured 3 subjects in different types of apparel (\textit{e.g.,} skirts and trousers).
Per subject, we captured around 20k multi-view video frames for training (only in-studio green background).
For testing, several separate 2k-frame videos in novel in-the-wild environments and poses are captured using a BlackMagic camera.
We recorded in different environments ({\em e.g.,} indoors, outdoors, in-studio) to test the generalization of our approach to novel lighting conditions. 
All the sequences include a large variety of different and challenging motions.
We apply a domain adaptation step proposed by Habermann et al.~\cite{habermann2020deepcap} by finetuning our pre-trained networks on the monocular test sequences.
For quantitative evaluations, we also recorded 5 in-studio sequences to be able to acquire ground truth meshes using multi-view stereo~\cite{metashape}.
\begin{figure}[t]
\centering
\includegraphics[width=0.85\textwidth]{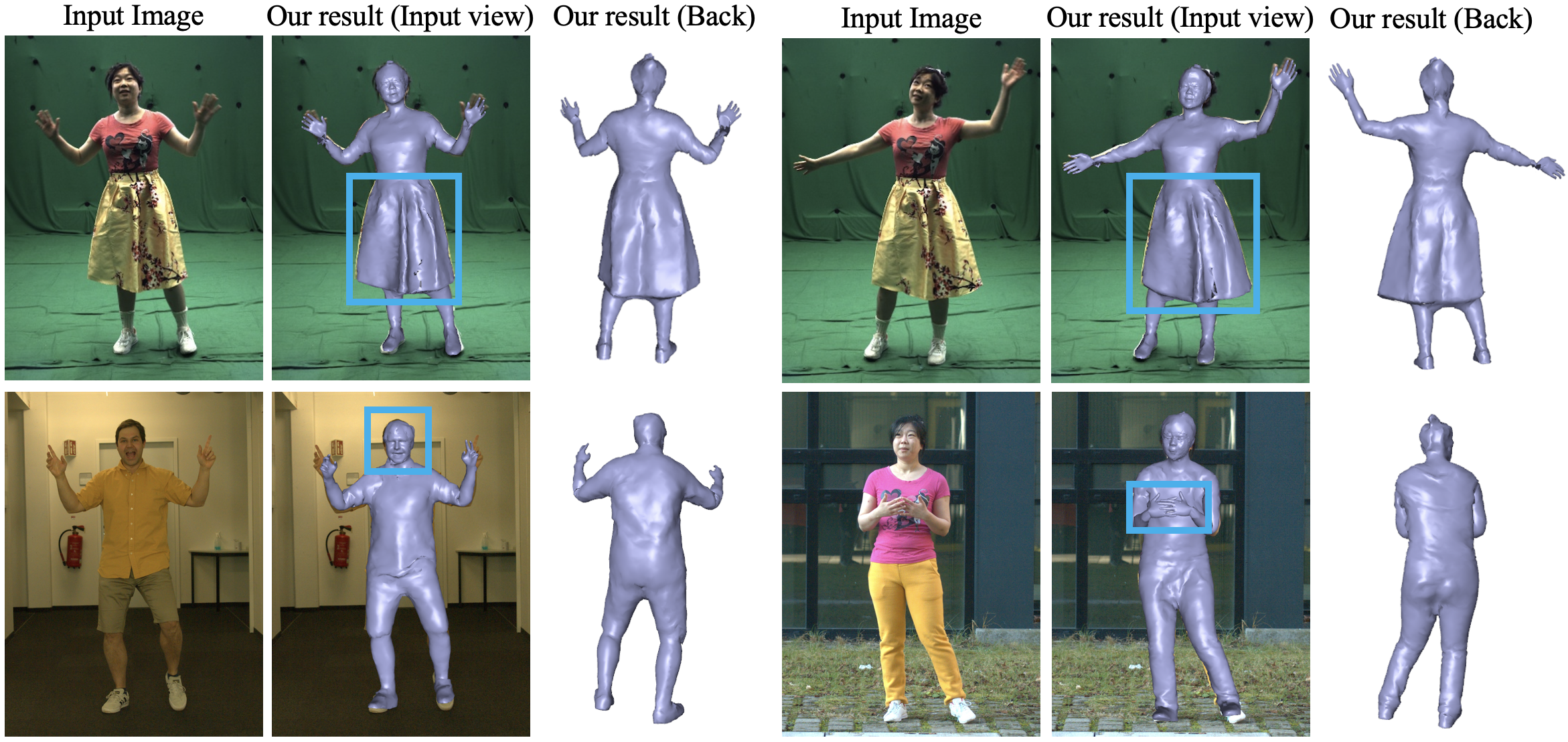}
  \caption{
  Qualitative results for subjects with different types of apparel, poses, and backgrounds. 
  Our method not only precisely overlays onto the input images, but also captures the wrinkle patterns nicely.
  Even the occluded regions look plausible in the back views.
  } 
  \label{fig:qualitative}
 \end{figure}
\begin{figure*}[!t]
\centering
\includegraphics[width=1.0\textwidth]{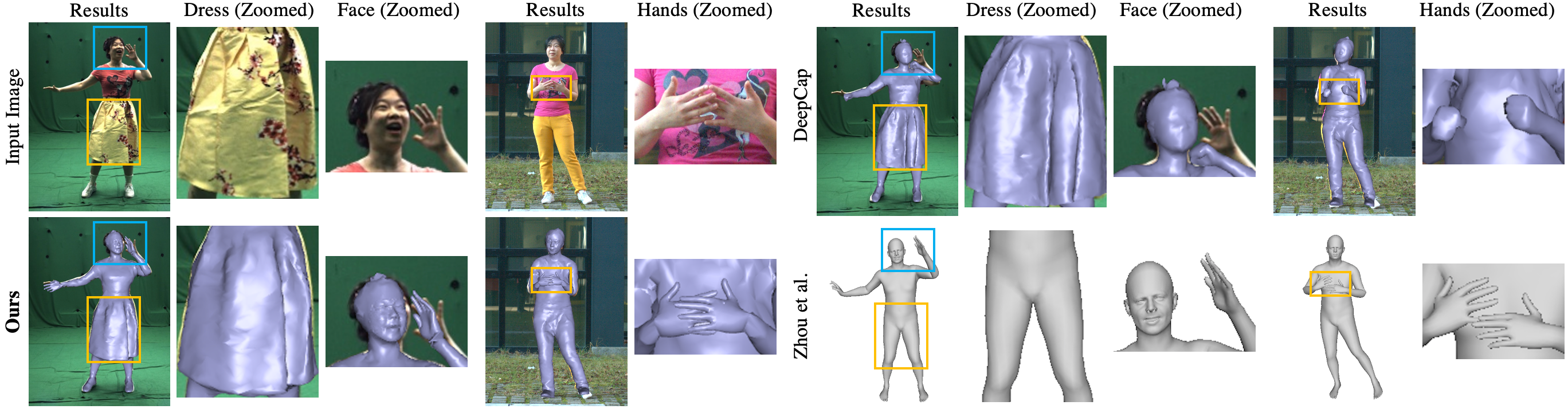}
  \caption{
  Comparisons. 
  Our method can capture high-frequency details on the non-rigid clothing surfaces, facial expressions, and hand gestures. 
  DeepCap~\cite{habermann2020deepcap} cannot dynamically capture face and hands, while Zhou et al.~\cite{zhou2021monocular} cannot capture the non-rigid deformation. 
  Our method can better capture dynamic non-rigid clothing details than DeepCap~\cite{habermann2020deepcap}.
} 
\vspace{-15pt}
  \label{fig:comparison}
 \end{figure*}
\begin{figure}[th!]
\centering
\includegraphics[width=0.8\textwidth]{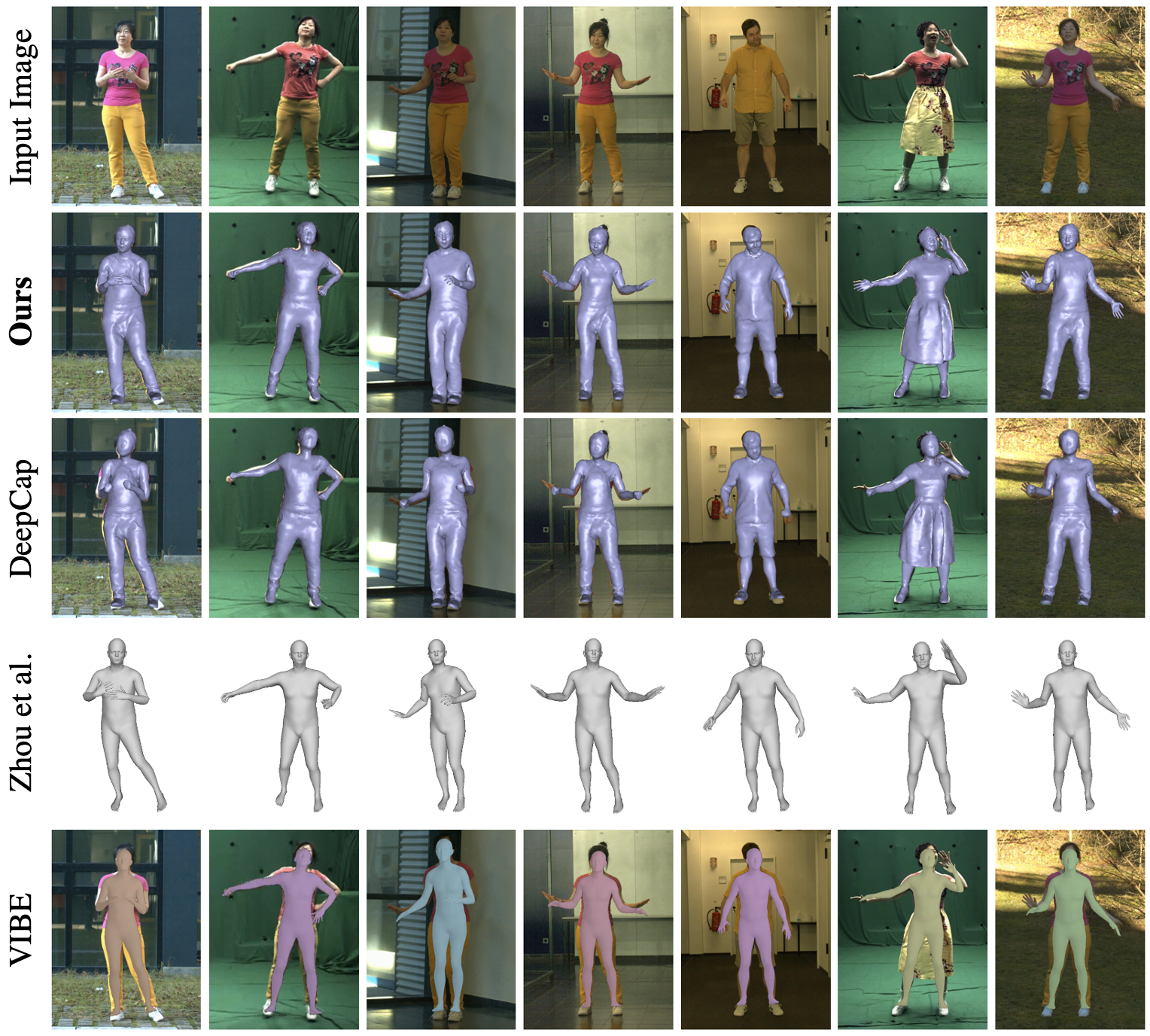}
\vspace{-8pt}
\caption{
Qualitative comparison.  Compared to other methods, HiFECap can better capture high-frequency details on the non-rigid clothing surfaces \textit{and} facial expressions as well as hand gestures and be generalizable to different clothing, motions, and backgrounds. 
}
\vspace{-12pt}
\label{fig:seq_comparison}
\end{figure}
\noindent\textbf{Qualitative Results.}
We visualize monocular results in Fig.~\ref{fig:qualitative} and the supplement with different clothing, motions, and backgrounds. 
Our reconstruction jointly captures facial expressions, hand poses, and high-frequency details on clothing.
It overlays precisely with the input images and achieves plausible results for the occluded areas. The recovered clothing wrinkles of the posed and deformed template match the ones in the input images.
%
%

%
\noindent\textbf{Comparisons.}
There is no dataset with joint ground-truth skeletal pose, hands, face, and cloth tracking and obtaining such is far from being trivial. So it is hard to quantitatively evaluate them in our setting. As an alternative, we show extensive qualitative results for face and hands in Fig.~\ref{fig:comparison} and our supplemental material.
In Fig.~\ref{fig:seq_comparison}, we further compare our results qualitatively with related approaches.
Compared to our approach, DeepCap~\cite{habermann2020deepcap} cannot capture high-frequency details on the clothing due to the limited capacity of the embedded graph and the silhouette-only supervision strategy.  
Our method can capture more accurate clothing details than DeepCap corresponding to the input video. Although DeepCap tracks the clothing, it outputs very different (mostly coarse and global) wrinkles compared to the ones observed in the input.
By leveraging image convolutions and graph convolutions, our new architecture, DisplaceNet, regresses the per-vertex displacement field. It captures the dynamic high-frequency details on the nonrigid deforming surface while 
DeepCap only deforms the static clothing of the input template by matching the silhouette of the deforming clothing using 2D supervision. 
Furthermore, DeepCap can only capture the pose and clothing deformations, while our approach can dynamically capture facial expressions and hand poses. 
Our input video captures the entire body without additional information about hands, and we further localize the hands to capture these parts. Thus, existing hand-only methods cannot be directly applied to our setting as our method requires cropping and alignments of the human's hands, face, and body. Concerning full-body methods, we show superior hands and face capture results compared to previous work.
Zhou et al.~\cite{zhou2019monocular} jointly regresses facial expressions, hand poses, and the body pose, but it is not able to capture the non-rigid deformation of the clothing at all, and VIBE~\cite{kocabas2020vibe} only captures body and hand poses.
%
%

\begin{figure*}[!t]
\centering
\includegraphics[width=0.9\textwidth]{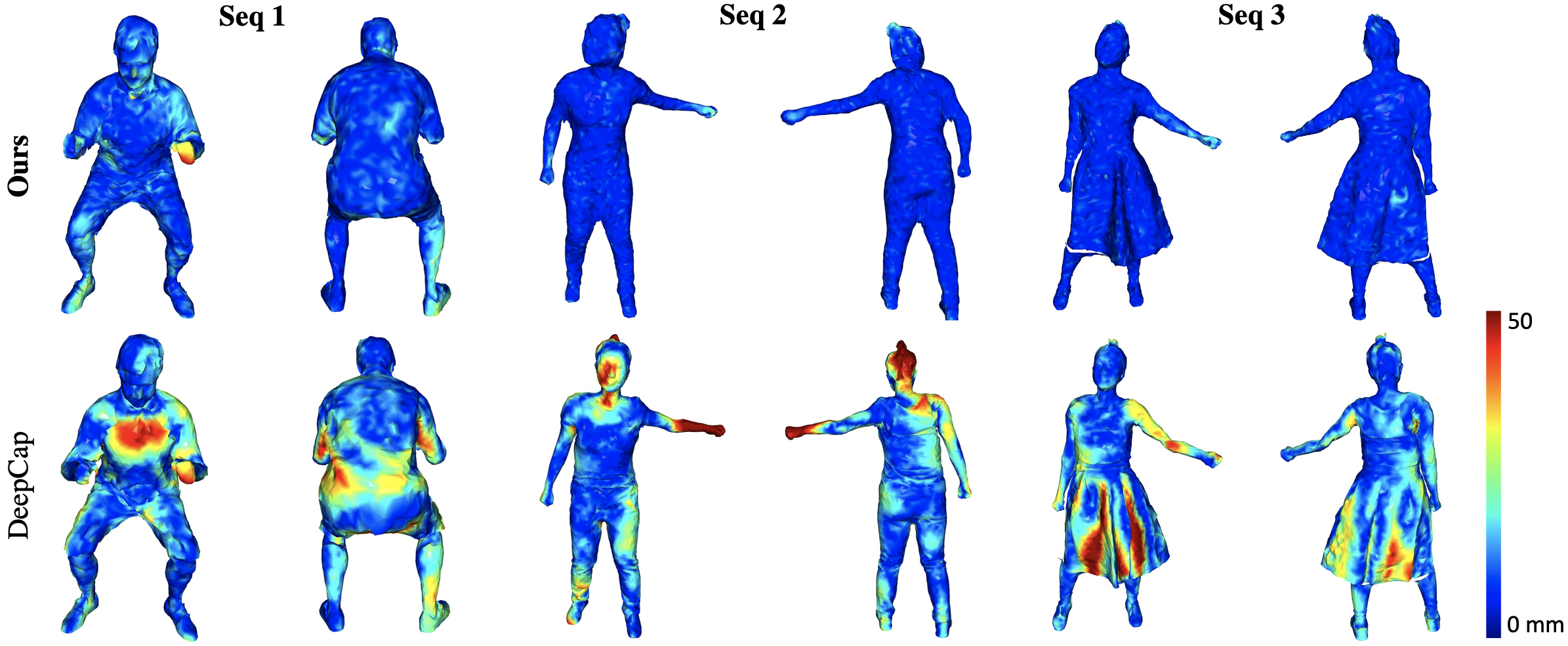}
\vspace{-10pt}
  \caption{
  Visualization of the per-vertex error (MSE) of our method and DeepCap~\cite{habermann2020deepcap} compared to the ground truth meshes.
  Our method has significantly lower error indicating our method better captures high-frequency non-rigid surface deformations. 
  } 
  \vspace{-10pt}
  \label{fig:comparison_deepcap}
 \end{figure*}

%
\noindent\textbf{Quantitative Results.}
We evaluate the accuracy of the recovered non-rigid deformation between our method and related works~\cite{habermann2020deepcap, zhou2021monocular, kocabas2020vibe}.
Tab.~\ref{tab:comparison} shows the quantitative results of these methods for three test sequences by computing the average Chamfer distance and the average symmetric Hausdorff distance between the output and ground truth meshes. 
We observe that our method achieves higher accuracy in terms of both metrics confirming that our reconstruction results can capture high-frequency details on the non-rigid parts.
In Fig.~\ref{fig:comparison_deepcap}, we show the 3D reconstruction results of our approach and the state-of-the-art approach DeepCap and visualize their per-vertex errors to the ground truth meshes. 
Note that we do not apply the dynamic hand pose and facial expressions here to evaluate the non-rigid clothing deformations separately.
Our method especially outperforms DeepCap in the dynamic clothing areas indicating that such dynamic deformations are better recovered by our approach.
%
%
%
\begin{table}[t]
	\begin{center}

    	\label{tab:comparison}
    	\resizebox{\textwidth}{!}{
    		\begin{tabular}{|c|c|c|c|c|c|c|}
    			\hline
    			&
    			\multicolumn{2}{|c|}{Sequence 1}  & 
    			\multicolumn{2}{|c|}{Sequence 2} & 
    			\multicolumn{2}{|c|}{Sequence 3}\\
    			\hline
    			\textbf{Method}                                 		& \textbf{Chamfer}$\downarrow$   	            & \textbf{Hausdorff}$\downarrow$   & \textbf{Chamfer}$\downarrow$   	            & \textbf{Hausdorff}$\downarrow$& \textbf{Chamfer}$\downarrow$   	            & \textbf{Hausdorff}$\downarrow$  \\
    			\hline
    			\textbf{Ours}							                & \textbf{7.25}	&	\textbf{39.72}                   & \textbf{9.21}   &      \textbf{40.07}        &        
    			\textbf{7.26} &   \textbf{32.69}       \\ 
    			\hline
    		DeepCap~\cite{habermann2020deepcap}  &
    		21.09 & 77.83  &
    		14.32 &  107.21   &
    		17.88 &  98.49 \\
    			\hline
    			Zhou et al~\cite{zhou2021monocular}  &
    	24.49	 & 133.95  &
    	51.35	 &  230.42   &
    	34.83	 &  157.11  \\
    			\hline
    			VIBE~\cite{kocabas2020vibe}  &
    	47.21	 & 121.15   &
    	72.00	 &  229.90   &
    	82.24	 & 224.60   \\
    			\hline
    		\end{tabular}
    	}
		
	\end{center}
	\vspace{-8pt}
	\caption
    	{
    		Quantitative comparisons. 
    		Our method significantly outperforms other approaches in terms of Chamfer distance and Hausdorff distance with respect to the ground truth. 
    		The accuracy of our method increased by almost 50\% compared to the state-of-the-art DeepCap. Our approach can better capture high-frequency details on the dynamic non-rigid surfaces.
    	}
    	\vspace{-15pt}
\end{table}
%
%
%
%
%
\vspace{-12pt}
\section{Conclusion} \label{sec:conclusion}
%
%
In this paper, we presented HiFECap, the first monocular human performance capture approach, which jointly tracks the body pose, hand gestures, facial expressions, and high-fidelity non-rigid surface deformations. 
We showed that higher-fidelity character surface tracking can be achieved by adding a dedicated displacement network to the character deformation process, which is a hybrid network architecture leveraging image convolutions and graph convolutions with locality preserving receptive fields.
Further, tightly coupling the template with parametric hand and face models enables the tracking of all aspects of the human.
In our experiments, we validated these design choices and show that we improve the current state-of-the-art in terms of space-time coherent surface tracking.
While this work is a clear step towards expressive capture of humans, we still believe that there is a lot of future work to be done, especially in the areas of physically correct human tracking and real-time performance, which would enable monocular performance capture in VR and AR settings.
\section{Acknowledgements} 

The authors from MPII were supported by the ERC Consolidator Grant 4DRepLy (770784). 
We would like to thank Mengyu Chu for her participation in our experiments and Zhi Li for the voice-over of our video.
\bibliography{egbib}

\appendix

\section{Appendix - Overview}
In the following, we show more qualitative results and visualize our undeformed template meshes (Sec.~\ref{supp:qual}), provide details about the network architectures (Sec.~\ref{supp:arch}), and give more explanations regarding the individual loss terms (Sec.~\ref{supp:loss}).
Last, we provide more details about the training process (Sec.~\ref{supp:train}) and limitations as well as future work (Sec.~\ref{supp:limitations}).
\begin{figure*}[th!]
\centering
\includegraphics[width=\textwidth]{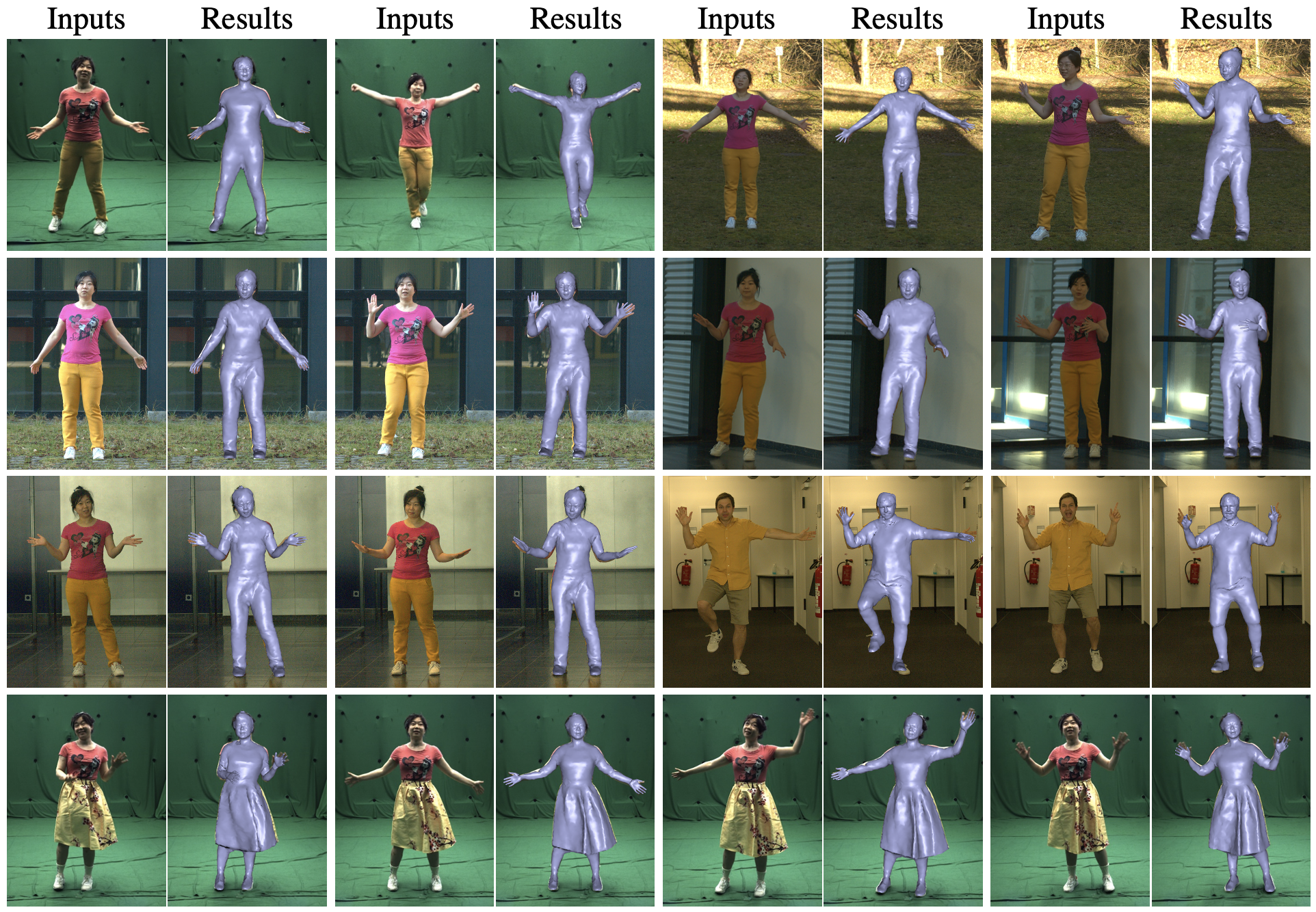}
  \caption{
  Qualitative results. 
  We show results for subjects with different types of apparel, poses, and backgrounds. 
  We show the overlay of our reconstruction onto the input frames.
  Note that our method not only precisely overlays onto the input images, but also captures the wrinkle patterns nicely.
  } 
  \label{fig:supp_qualitative}
 \end{figure*}

\section{Qualitative Results and Template Meshes} \label{supp:qual}

We visualize some additional qualitative results in Fig.~\ref{fig:supp_qualitative}, which include different actors, clothing styles, body motions, backgrounds, facial expressions, and hand gestures.
The results show that our reconstruction can jointly capture facial expressions, hand poses, and also high-frequency details, such as deforming wrinkles on the clothes.
\begin{figure}[t]

\centering

\includegraphics[width=\textwidth]{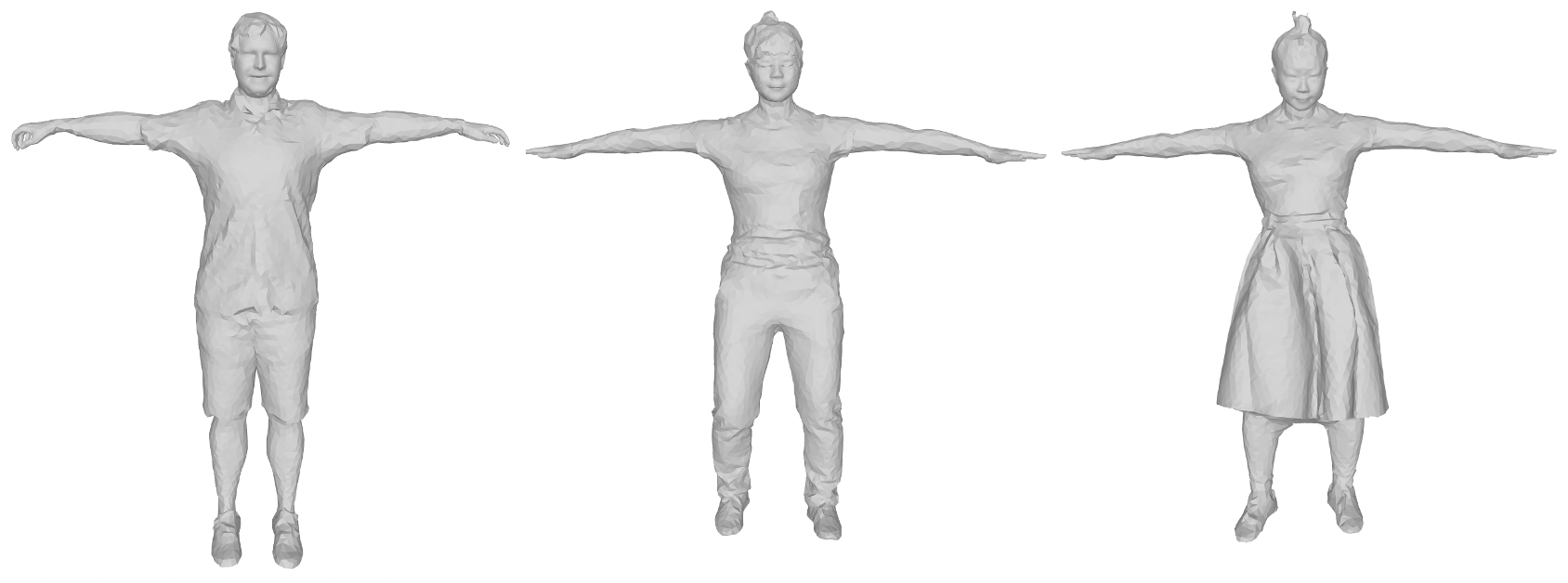}

\caption{
Examples of the final template meshes with hands and face.
}
\label{fig:final_templates}
  
\end{figure}
We show our final templates after further adding hands and face in Fig.~\ref{fig:final_templates}.

\section{Ablation Study} \label{supp:arch}

\begin{figure}[th!]

\centering

\includegraphics[width=\textwidth]{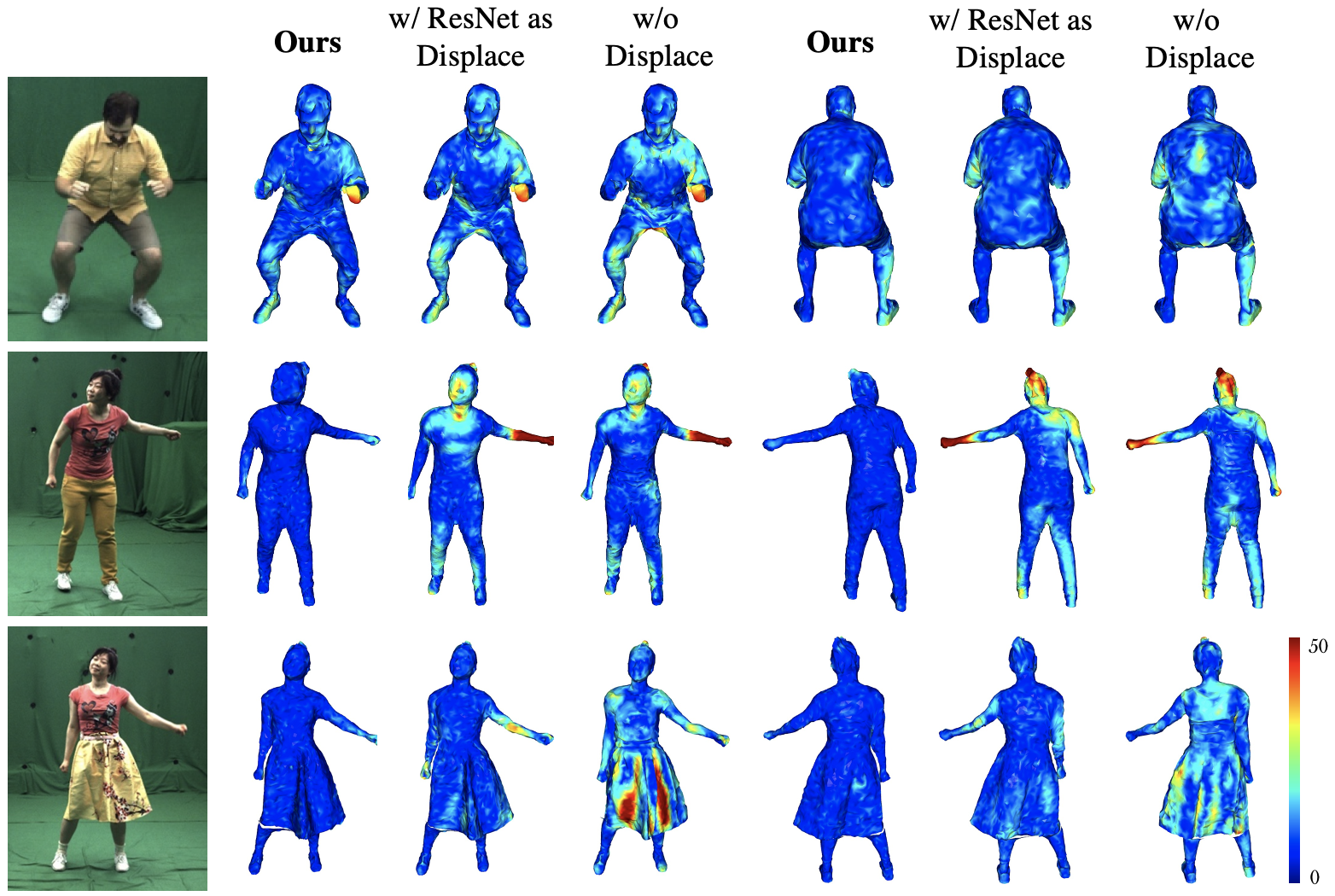}

\caption{Ablation studies. 
Visualization of the per-vertex error (MSE) of our method, our method without $DisplaceNet$, and another baseline where we replace our proposed architecture with a pure image-convolutional one.
Our method has significantly lower error than the baselines, which validates our design choices.
  } 
\label{fig:supp_ablation}

\end{figure}

We propose a combined image- and graph-convolutional architecture, called $DisplaceNet$, to regress the per-vertex displacement field, which captures the high-frequency details on the nonrigid deforming surface.
In Fig.~\ref{fig:supp_ablation}, we compare our method to a purely image-based convolutional architecture~\cite{he2016residual} and a baseline, which does not use any dense per-vertex displacements.
The comparison results demonstrate that our design achieves the best result as indicated by the per-vertex error and recovers richer surface details such as the wrinkles of the clothing.
Note that especially in those regions, the baselines fall short in recovering the geometric details.
Thus, the highest error for the baselines can be observed in these regions while our method shows a significantly lower error.

\begin{table}
 	\label{tab:ablation}	
	\begin{center}

    	\resizebox{\textwidth}{!}{
    		\begin{tabular}{|c|c|c|c|c|c|c|c|c|}
    			\hline
    		\multicolumn{3}{|c|}{\textbf{Method}}  	&
    			\multicolumn{2}{|c|}{Sequence 1}  & 
    			\multicolumn{2}{|c|}{Sequence 2} & 
    			\multicolumn{2}{|c|}{Sequence 3}\\
    			\hline
    				Rend.							                & Chamfer	& Displ.  
    			                               		& \textbf{Chamfer}$\downarrow$   	            & \textbf{Hausd.}$\downarrow$   & \textbf{Chamfer}$\downarrow$   	            & \textbf{Hausd.}$\downarrow$& \textbf{Chamfer}$\downarrow$   	            & \textbf{Hausd.}$\downarrow$  \\
    			\hline
    			\greencheck & \greencheck  & \greencheck							                & \textbf{7.25}	&	\textbf{39.72}                  & \textbf{9.21}   &      \textbf{40.07}       &        
    			\textbf{7.26} &   \textbf{32.69}     \\ 
    			\hline
    		\greencheck & \greencheck  & \textbf{\color{blue}{+}} &
    			9.17 & 41.82  &
    			9.62 &  51.70  &
    			8.75 &  38.91  \\
    			\hline
    			\redcross & \greencheck & \greencheck &
    			15.66 & 51.21   &
    			10.88 &  71.53 &
    			9.80 &  46.82   \\
    			\hline
    			\greencheck & \redcross & \greencheck &
    			18.21 & 65.45 & 
    			 12.23 &  93.64  &
    			15.43&  89.77      \\
    			\hline
    			\greencheck & \greencheck & \redcross  &
    			18.34 & 65.85 &
    			 12.70 &  88.78  &
    			16.32 &   75.47    \\
    			\hline
    			\redcross & \redcross & \greencheck &
    			18.47 & 72.81 &
    			 12.92 & 99.16  &
    			16.73 &   93.58    \\
    			\hline
    			\redcross & \greencheck & \redcross &
    			18.76 & 69.19 &
    			 13.17 & 97.79  &
    			17.52 &   80.71    \\
    			\hline
    			\greencheck & \redcross & \redcross &
    			20.15 & 75.46 &
    			 13.45 & 105.62  &
    			17.81 &   96.71    \\
    			\hline
    			\redcross & \redcross & \redcross  &
    		21.09 & 77.83   &
    		14.32 &  107.21    &
    		17.88 &  98.49    \\
    			\hline

    		\end{tabular}
    	}
		
	\end{center}
	\caption
    	{
    		Quantitative ablation study. 
    		We evaluate the necessity of the different loss terms and the visibility-aware and rigidity-aware displacement network in terms of Chamfer distance and Hausdorff distance compared to the ground truth meshes. 
    		``\textbf{\color{blue}{+}}'' indicates that we use a ResNet architecture instead of our proposed \textit{DisplaceNet} architecture.
            Note that our specific design choices provide the best results in terms of surface tracking accuracy compared to the baselines.
    	}
\end{table}
Next, in Tab.~2, we evaluate our design choices concerning the novel displacement network architecture and the proposed supervision strategy.
To this end, we first replace the \textit{DisplaceNet} architecture with a ResNet50 that has a fully connected backbone predicting the displacement field directly (second result row). 
The result shows that our proposed architecture improves the quality of deformation.
This is due to the fact, that the receptive field of our proposed architecture is more local and, thus, the deep features attached to the graph are better suited for predicting these local displacement vectors compared to the ResNet50 architecture, which in its last fully connected layer creates a fully global dependency between the spatial image features.
%
%
\par
\begin{figure}[!ht]
\centering
\includegraphics[width=0.7\linewidth]{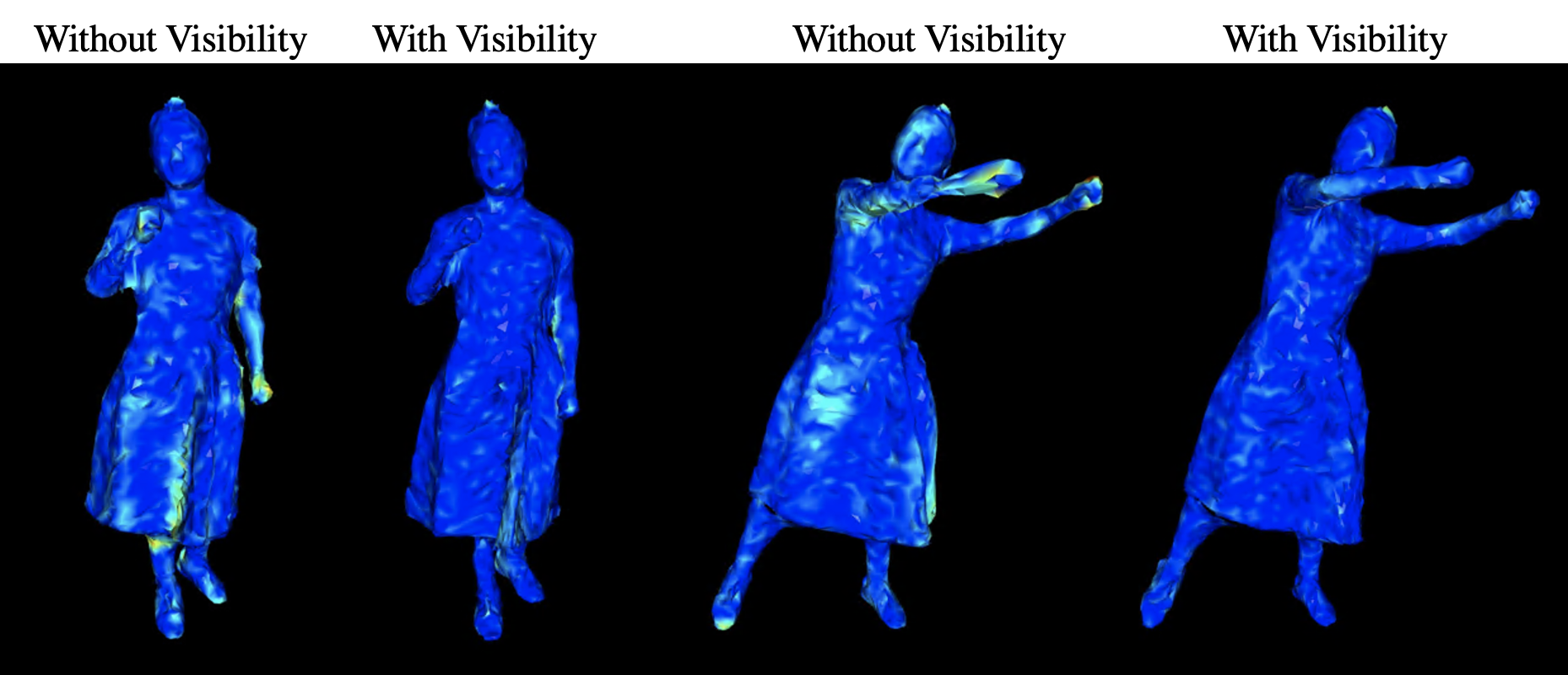}
  \caption{Error comparison showing how the visibility-modulated feature map improves the overall reconstruction performance.
} 
  \label{fig:visibility}

 \end{figure}

 
To show the importance of the different loss terms and the proposed visibility-aware and rigidity-aware vertex displacement network \textit{DisplaceNet}, we conduct an ablation study where we remove each of these components from the non-rigid training process. 
We report the average Chamfer distance and the average symmetric Hausdorff distance and visualize their per-vertex errors. 
We can see that removing any component from our carefully designed combination reduces the quality of the estimated deformations, which confirms that our design choice indeed provides the best performance.
Furthermore, we show an error comparison of how the visibility-modulated feature map improves the overall reconstruction performance in Fig.~\ref{fig:visibility}.

\section{Details about the Loss Terms} \label{supp:loss}
%
%
\subsection{Silhouette Loss} 
We define the multi-view silhouette loss for the input frame $f$ as 
\begin{equation} 
\mathcal{L}_{\textrm{sil}}(\mathbf{V}) = \sum_{c=1}^C\sum_{i\in\mathcal{B}_c} d_{c, i} \norm{\mathcal{D}_{c} (\pi_c (\mathbf{V}_i))}^2. 
\end{equation} 
The silhouette loss $\mathcal{L}_{\textrm{sil}}$ encourages the deformed mesh to fit the multi-view image silhouettes from all the cameras.
$\mathbf{V}_i$ is the position of the $i$-th vertex of the posed and the deformed template mesh obtained by the deformation layer. 
$\pi_c$, the perspective camera projection of the camera $c$, projects $\mathbf{V}_i$ onto the image plane. 
$d_{c, i}\in\{-1, +1\}$ is a directional weight to encourage the optimization to follow the right direction in the distance field~\cite{habermann2019livecap} and $\mathcal{B}_c$ is the set of boundary vertices computed by rendering the depth maps and detecting whether the projected vertex $\pi_c (\mathbf{V}_i)$ is near the intersection of background and foreground pixels. 
Here, $\mathcal{D}_{c}$ is the distance transform image of camera $c$.
%
%
\subsection{Marker Loss} 
Similar to the 2D keypoint loss for \textit{PoseNet}, we define the marker loss as the weighted squared loss of the difference between the projected 3D marker on the posed mesh from the camera $c$ and the detected 2D marker on the input image.
The loss is defines as 
\begin{equation}
\mathcal{L}_{\textrm{mk}} = \sum_c \sum_m \beta_{c, j}\norm{\pi_c (\mathbf{M}_{j}) - \mathbf{m}_{c, j}}^2, 
\end{equation}
where $\pi_c (\mathbf{M}_{j})$ is the projected 3D marker $j$ on the posed mesh from the camera $c$, $\mathbf{m}_{c, j}$ is the detected 2D marker on the input image, and $\beta_{c, j}$ is its corresponding confidence value. 
%
%
\subsection{As-Rigid-As-Possible Deformation Loss} 
To ensure the smoothness of the mesh surface, we apply as-rigid-as-possible deformation loss $\mathcal{L}_{\textrm{arap}}$~\cite{sorkine2007arap} and use material-aware weights~\cite{habermann2019livecap} to define the different levels of rigidity for embedded graph nodes ({\em e.g.,} clothing is assigned to a lower rigidity weight than the skin, and, thus, it has more freedom to deform than skin). 
%
%
\subsection{Differentiable Rendering Loss} 
The multi-view silhouette-based loss can be used to get the posed and coarsely deformed mesh. 
However, it does not allow to capture fine non-rigid deformations such as surface folds. 
Thus, we deploy a dense rendering loss 
\begin{equation}
\begin{split}
\mathcal{L}_{\textrm{dr}}(\mathbf{V}, L) = \sum_c \norm{R_c(\mathbf{V}, L_c, \mathcal{T}) - \mathcal{I}_c}^2 
\end{split},
\end{equation}
which takes the posed and deformed mesh and the static texture, renders it from various camera views, and compares the rendered images $R_c(\mathbf{V}, L, \mathcal{T})$ under the camera's lighting condition $L$ with the corresponding input frame $\mathcal{I}_c$: 
To deal with different light conditions, camera optics and scene reflections, we optimize the lighting parameters for all the cameras used in the multi-camera sequences. 
Under the assumption of Lambertian material and smooth lighting environment, we apply the spherical harmonics (\textit{SH}) lighting representation~\cite{muller2006spherical} to model the lighting condition $L_c$ of each camera based on $27$ lighting coefficients $\mathbf{l}_c\in\mathbb{R}^{9\times3}$.
There are nine SH basis functions and for each SH basis function $\textit{SH}_j$, and we have $\mathbf{l}_{c, j}\in\mathbb{R}^{3}$ for each color channel. 
Then, the lighting condition for each camera $L_c(\mathbf{V}, \mathbf{l}_c)$ can be computed as 
\begin{equation}
\begin{split}
L_c(\mathbf{V}, \mathbf{l}_c) = \sum_{j=1}^9 \mathbf{l}_{c,j}\textit{SH}_j(n_c(\mathbf{V})), 
\end{split}
\end{equation}
where $n_c(\mathbf{V})$ represents the image pixel normal based on the underlying geometry. 
Then, the rendering function can be defined as 
\begin{equation}
\begin{split}
R_c(\mathbf{V}, \mathbf{l}_c, \mathcal{T}) = A_c(\mathbf{V}, \mathcal{T}) \cdot L_c(\mathbf{V}, \mathbf{l}_c)
\end{split}.
\end{equation}
Given the vertex positions $\mathbf{V}$, the lighting coefficients $\mathbf{l}_c$ the texture $\mathcal{T}$---and under the assumption of Lambertian surface---the rendering function equals to the dot product of the albedo of the projected surface $A_c(\mathbf{V}, \mathcal{T})$ and the lighting condition $L_c(\mathbf{V}, \mathbf{l}_c)$. 
The optimized lighting condition $\mathbf{l}_c$ for each camera $c$ is then 
\begin{equation}
\begin{split}
\mathop{\mathrm{argmin}}\nolimits_{\mathbf{l}_c}\norm{R_c(\mathbf{V}, \mathbf{l}_c, \mathcal{T}) - \mathcal{I}_c}^2. 
\end{split}
\end{equation}
To optimize the lighting coefficients across all the frames in the training sequence, we apply the Adam optimizer~\cite{Kingma2015AdamAM} to minimize $\norm{R_c(\mathbf{V}, \mathbf{l}_c, \mathcal{T}) - \mathcal{I}_c}^2$ by iteratively sampling the training frames. 
%
%
\subsection{Isometry Loss}
We employ an Isometry loss 
\begin{equation}
\begin{split}
\mathcal{L}_{\textrm{iso}}(\mathbf{V}) = \sum_{i}\sum_{(i, j)\in E} \dfrac{\textit{r}(\mathbf{V}_i, \mathbf{V}_j)}{\abs{(i, j)\in E}}
\abs{\norm{\mathbf{V}_i - \mathbf{V}_j} - \norm{\mathbf{V}_{\textrm{Relax}, i} - \mathbf{V}_{\textrm{Relax}, j}}}^2
\end{split}
\end{equation}
on the mesh geometry to encourage consistent edge lengths with respect to the undeformed template mesh. 
$E$ represents all the edges on the mesh,
$\mathbf{V}_i$ is the vertex position of the $i$-th vertex on the posed and deformed mesh and $\mathbf{V}_j$ where $(i, j)\in E$, is any vertex connecting to $\mathbf{V}_i$. 
$\mathbf{V}_{\textrm{Relax}, i}$ and $\mathbf{V}_{\textrm{Relax}, j}$ are their corresponding vertices on the original (unposed and undeformed) template mesh.
$\textit{r}(\mathbf{V}_i, \mathbf{V}_j)$ are the predefined per-edge rigidity weights based on the rigidity of body parts similar to the per-vertex rigidity weights of the template. 
Lower rigidity weights are assigned to materials with less non-rigid deformations.
For example, the skin has less deformation ability, so we assign weight 200.0 to the face and 50.0 to other parts of the skin. 
For more deformable parts, we assign lower weights, {\em e.g.,} we assign weight 1.0 to dresses, 2.0 to upper clothes, and 2.5 to pants. 
The isometry loss encourages that the length of every edge in the posed and deformed mesh is similar to the corresponding edge in the original unposed and undeformed mesh to penalize large stretching of the surface. 
%
%
\subsection{Laplacian Loss}
To avoid geometry distortion, we regularize the mesh with a Laplacian regularization term
\begin{equation}
\begin{split}
\mathcal{L}_{\textrm{lap}}(V) = \sum_i w_i\norm{\abs{(i, j)\in E}(\mathbf{V}_i - \mathbf{V}_{\textrm{Relax}, i}) - \sum_{(i, j)\in E} (\mathbf{V}_j - \mathbf{V}_{\textrm{Relax}, j})}^2
\end{split}
\end{equation}
where $w_i$ represents the spatially varying regularization weights for each vertex on the mesh. 
The Laplacian loss term ensures that adding the vertex displacements does not largely change the Laplacian of the mesh so that the mesh has a smooth surface.
\subsection{Tracking of Hands and Face}
\begin{figure}[t]

\centering

\includegraphics[width=\textwidth]{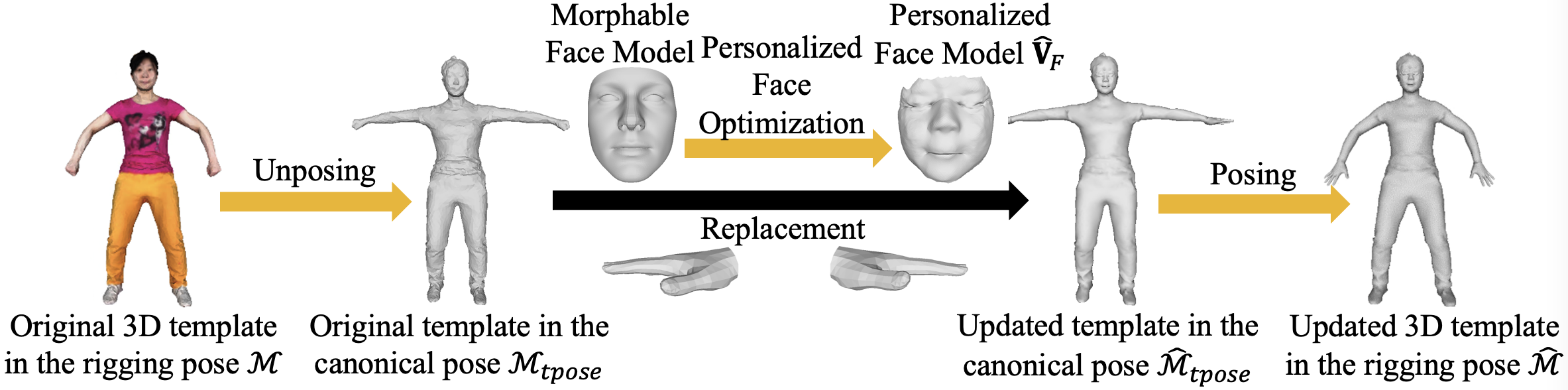}

\caption{
We apply Dual Quaternion Skinning to unpose the original template mesh to T-pose.
We then replace the face and hands of the original template with the optimized models to obtain the updated template mesh that contains the controllable hand and face models.
}
\label{fig:template}
  
\end{figure}

Fig.~\ref{fig:template} shows the process of replacing the face and hand regions on the template mesh with a parametric 3D face model~\cite{blanz1999morphable} and hand model~\cite{romero2017mano} to enable the joint tracking of hands and face.

\section{Training Details} \label{supp:train}

We train and test our approach on an NVIDIA Quadro RTX8000 GPUs with 48GB of memory.
Our training process includes 4 different stages, {\em i.e.,} 
\begin{enumerate} 
\item {Training \textit{PoseNet} with 2D keypoint Loss;}
\item{Training \textit{EDefNet} with silhouette loss and the regularization terms;}
\item{Training \textit{EDefNet} with all the loss terms;}
\item{Training \textit{DisplaceNet} with all the loss terms.}
\end{enumerate} 
We use the Adam optimizer~\cite{Kingma2015AdamAM} for all the training stages. 
All the network architectures are implemented in the Tensorflow framework. 
We train each stage for 120k iterations with a learning rate of $10^{-5}$. 
The training data are multi-view videos captured in studio with around 100 cameras.
Due to the limited memory and training time, instead of training on all the multi-view inputs, for each iteration, we randomly sample 30 camera views for all the multi-view loss terms.
The entire training process takes about 4 days in total if we train on 4 GPUs in parallel with a batch size of 8. 
At training time, the system requires multi-view videos, however, at test time, it only takes single-view videos. 
Thus, at test time, our method takes about 0.3 seconds on a single GPU.
%
%

%
%
\section{Discussion and Future Work} \label{supp:limitations}
%
%
Although our approach is the first to jointly track the human pose, facial expressions, hand gestures, and non-rigid clothing solely from a monocular video, it still has some limitations that open up future work in this direction.
Hands and face capture still has some room to improve. Our input video captures the entire body without any additional information about the hands and face, and we further localize the hands and face to capture these parts. Thus, existing face-only and hand-only methods cannot be directly applied to our setting as our method requires cropping and alignments of the human's hands, face, and body. Concerning full-body methods, we show the superior performance of hands and face capture results compared to previous work as shown in our qualitative results (Zhou et al.~\cite{zhou2021monocular} and VIBE~\cite{kocabas2020vibe}). It is hard to compare face and hand results quantitatively because it is very hard to obtain the ground truth meshes under this setting ({\em i.e.,} marker-less multi-view full-body capture).
Also, extreme body poses and hand gestures that are too different from the training motions can lead to erroneous pose estimation and deformation. 
To tackle this problem, we envision that in the future the overall robustness of our method can be further improved by jointly training on, both, in-studio multi-view data and in-the-wild monocular data. 
Moreover, tighter integration of physics into the capture process could also be interesting, i.e. physically more accurate skeletal pose estimation and surface deformations.
For now, our method is a per-frame approach and we mainly focused on the expressiveness of our output, {\em i.e.,} capturing all aspects of the human. However, there are many more directions to explore in this field, one of them being the temporal aspect. Thus, future work could involve modeling the temporal domain more explicitly in the neural network architecture.
Finally, we believe that our method has the potential to run in real-time, but for this, the foreground segmentation and 2D keypoint detection have to be more tightly linked to the regression of the character parameters, i.e. skeletal pose, surface deformations, hand gestures, and facial expressions.

\end{document}